\documentclass{article}

\usepackage{microtype}
\usepackage{graphicx}
\usepackage{subcaption}
\usepackage{booktabs} % for professional tables

\usepackage{hyperref}

\usepackage{xurl}

\usepackage[preprint]{icml2026}

\usepackage{amsmath}
\usepackage{amssymb}
\usepackage{mathtools}
\usepackage{amsthm}

\usepackage{cancel}
\usepackage{multirow}
\usepackage[table,xcdraw]{xcolor}
\usepackage{makecell}
\usepackage{xcolor}
\usepackage{pifont}
\usepackage{stfloats}
\usepackage{bbm}
\usepackage{enumitem}

\usepackage[most]{tcolorbox}

\usepackage[capitalize,noabbrev]{cleveref}

\theoremstyle{plain}

\theoremstyle{definition}

\theoremstyle{remark}

\usepackage[textsize=tiny]{todonotes}

\icmltitlerunning{CUDABench: Benchmarking LLMs for Text-to-CUDA Generation}

\begin{document}

% \twocolumn[
%   \icmltitle{Submission and Formatting Instructions for \\
%     International Conference on Machine Learning (ICML 2026)}
\twocolumn[
  \icmltitle{CUDABench: Benchmarking LLMs for Text-to-CUDA Generation}



  \icmlsetsymbol{equal}{*}

  \begin{icmlauthorlist}
    \icmlauthor{Jiace Zhu}{yyy}
    \icmlauthor{Wentao Chen}{yyy}
    \icmlauthor{Qi Fan}{yyy}
    \icmlauthor{Zhixing Ren}{yyy}
    \icmlauthor{Junying Wu}{yyy}
    \icmlauthor{Xing Zhe Chai}{yyy}
    \icmlauthor{Chotiwit Rungrueangwutthinon}{yyy}
    \icmlauthor{Yehan Ma}{yyy}
    \icmlauthor{An Zou}{yyy}
  \end{icmlauthorlist}

% Jiace Zhu, Wentao Chen, Qi Fan, Zhixing Ren, Junying Wu, Xing Zhe Chai, Chotiwit Rungrueangwutthinon, Yehan Ma, An Zou

  \icmlaffiliation{yyy}{Shanghai Jiao Tong University, Shanghai, China}

  \icmlcorrespondingauthor{An Zou}{an.zou@sjtu.edu.cn}

  \icmlkeywords{Machine Learning, ICML}

  \vskip 0.3in
]



\printAffiliationsAndNotice{}  % no special notice (required even if empty)

\begin{abstract}
Recent studies have demonstrated the potential of Large Language Models (LLMs) in generating GPU Kernels. Current benchmarks focus on the translation of high-level languages into CUDA, overlooking the more general and challenging task of text-to-CUDA generation. Furthermore, given the hardware-specific and performance-critical features of GPU programming, accurately assessing the performance of LLM-generated GPU programs is nontrivial. In this work, we introduce \textbf{CUDABench}, a comprehensive benchmark designed to evaluate the text-to-CUDA capabilities of LLMs. First, we construct \textbf{CUDABench-Set}, which covers Breadth-Depth-Difficulty evaluation space in diverse application domains, including artificial intelligence, scientific computing, and data analytics, etc. Furthermore, we propose \textbf{CUDABench-Score} and \textbf{Generative Verification Pipeline} that assess (1) compilation correctness, (2) functional consistency through execution-based verification, and (3) a novel roofline-based metric, Performance-Score. Benchmarking state-of-the-art LLMs reveals insightful findings and challenges of text-to-CUDA, such as a notable mismatch between high compilation success rates and low functional correctness, a lack of domain-specific algorithmic knowledge, and suboptimal utilization of GPU hardware resources. Our benchmark is available at \url{https://github.com/CUDA-Bench/CUDABench}.
\end{abstract}

\section{Introduction}

The rapid advancement of Large Language Models (LLMs) has gained increasing interest in their application to code generation~\cite{chen2021evaluating}, including GPU programming. Recent studies demonstrate that LLMs are capable of generating correct and executable GPU programs in CUDA \cite{ouyang2025kernelbench,chen2025cudallm}, thereby lowering the barrier for developers to leverage the computational power of modern GPUs.
Despite these promising early results, a comprehensive benchmarking of LLMs for GPU program generation remains limited. Existing benchmarks and evaluations focus primarily on the translation of high-level programming languages, such as PyTorch, into CUDA \cite{ouyang2025kernelbench,li2025tritonbench,wen2022babeltower}. 
This setting evaluates LLMs in a code-to-code translation regime, where the program structure and computational intent are explicitly specified in the input code.
As a result, such benchmarks fail to capture the more general and challenging scenario of text-to-CUDA generation, which requires covering the breadth of diverse domains beyond native PyTorch Machine Learning (ML) workloads (e.g., Scientific Simulation) at production-level input scales, while forcing LLMs to infer algorithmic intent and implementation details directly from natural language.

Moreover, CUDA kernel is inherently hardware-specific and performance-critical~\cite{kirk2016programming,ryoo2008optimization}. Writing a high-performance CUDA kernel requires careful consideration of architectural details, specifically balancing computational throughput and memory bandwidth to improve performance facing both compute and memory bounds~\cite{guide2020cuda}. Consequently, evaluating LLM-generated kernels solely based on functional correctness is insufficient. A kernel that produces correct output may still suffer from severe performance degradation compared to optimized implementations~\cite{kirk2016programming,ryoo2008optimization}. 
Therefore, accurately assessing kernel performance and understanding how close it comes to expert-level performance are nontrivial.

\begin{figure*}[t]
\centering
\includegraphics[width=0.95\textwidth]{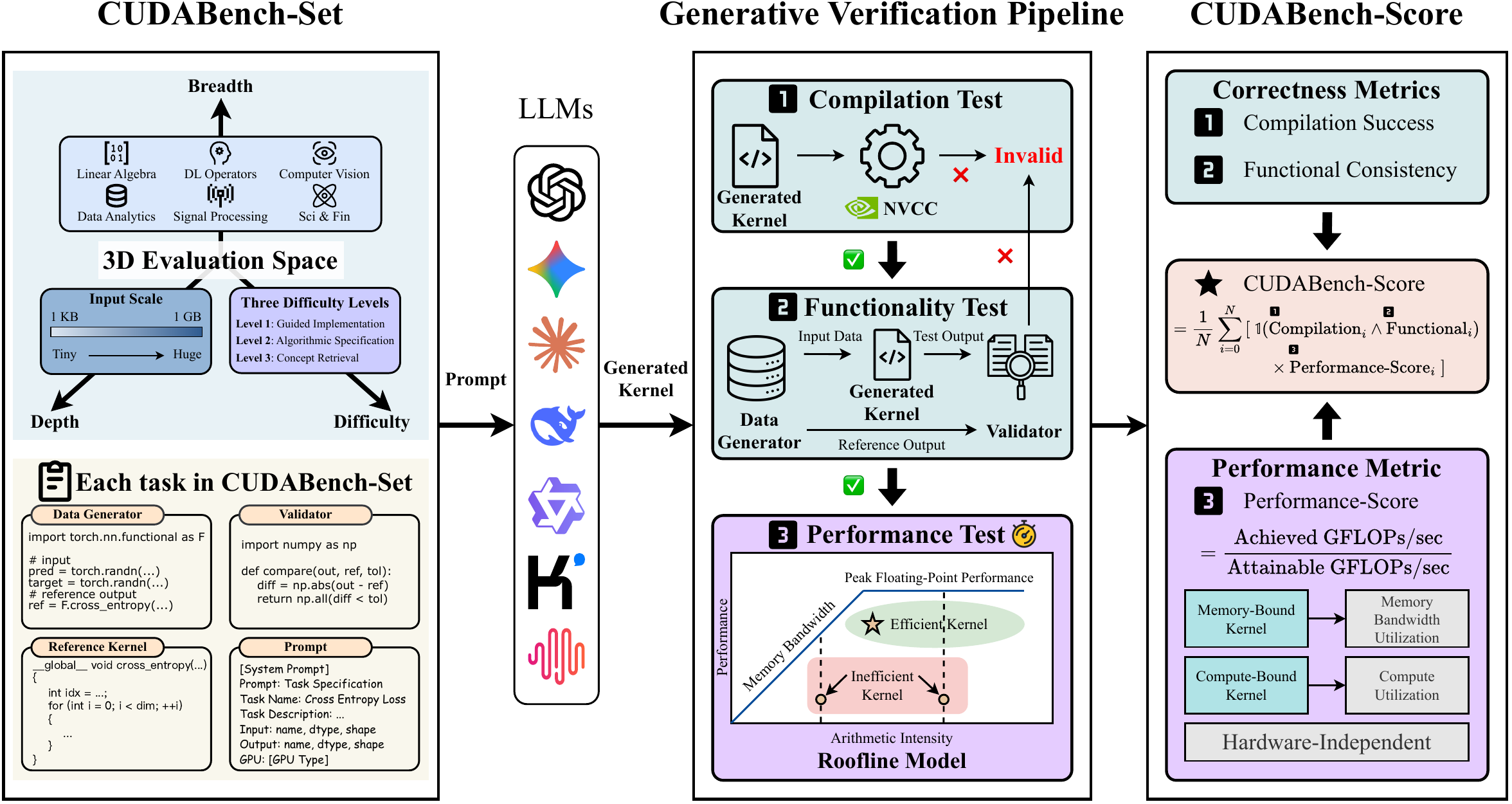}
\caption{Overview of CUDABench.}
\label{fig:overview}
\end{figure*}

These challenges highlight the necessity for benchmarks and evaluation methodologies that (1) target text-to-CUDA generation directly, (2) provide coverage across multiple application domains, and (3) incorporate hardware-independent metrics that reflect the practical performance of GPU programming.
Therefore, in this work, we present \textbf{CUDABench} (as shown in Fig. \ref{fig:overview} and Tab. \ref{tab:compare}), a comprehensive benchmark to systematically evaluate the text-to-CUDA program generation capabilities of LLMs across diverse domains, including but not limited to PyTorch ML workloads kernels. The main contributions are as follows.

\begin{itemize}[leftmargin=*]
    \item We construct \textbf{CUDABench-Set}, a text-to-CUDA dataset that covers a Breadth-Depth-Difficulty three-dimensional evaluation space, encompassing multiple application domains, different input scales, and varying difficulty levels of prompt details.
    \item We propose \textbf{Generative Verification Pipeline}, which evaluates compilation and functional correctness through strict execution-based verification, and \textbf{CUDABench-Score}, a novel hardware-independent metric that measures the comprehensive performance of generated kernels beyond execution time. 
    \item We conduct extensive evaluation of state-of-the-art LLMs using CUDABench, revealing their strengths, limitations, and several insightful findings about LLM behaviors when generating CUDA kernels.
\end{itemize}

% \usepackage{booktabs}
% \usepackage{multirow}
\begin{table*}[]
\centering
\caption{Difference between CUDABench and existing benchmarks.}
\label{tab:compare}
\resizebox{\textwidth}{!}{
\begin{tabular}{@{}llccccccc@{}}
\toprule
\multirow{2.5}{*}{\textbf{\makecell{Kernel Generation\\Benchmark}}} & 
\multirow{2.5}{*}{\textbf{Paradigm}} & 
\multicolumn{2}{c}{\textbf{Application Domain Coverage}} & 
\multicolumn{2}{c}{\textbf{Evaluation Infrastructure}} & 
\multicolumn{3}{c}{\textbf{Evaluation Metrics}} \\ 
\cmidrule(lr){3-4} \cmidrule(lr){5-6} \cmidrule(lr){7-9}
 &  & 
\textbf{\makecell{Machine Learning\\Workloads}} & 
\textbf{\makecell{Holistic Domain\\Coverage}} & 
\textbf{\makecell{Reference\\Implementations}} & 
\textbf{\makecell{Realistic\\Input Sizes}} & 
\textbf{Correctness} & 
\textbf{Execution Time} & 
\textbf{\makecell{Memory Bandwidth \&\\Compute Utilization}} \\ 
\midrule
KernelBench \cite{ouyang2025kernelbench} & PyTorch-to-CUDA & \textcolor{green!60!black}{\ding{51}}      & \textcolor{red}{\ding{55}} & \textcolor{red}{\ding{55}} & \textcolor{red}{\ding{55}} & \textcolor{green!60!black}{\ding{51}} & \textcolor{green!60!black}{\ding{51}} & \textcolor{red}{\ding{55}} \\
TritonBench \cite{li2025tritonbench} & PyTorch-to-Triton &  \textcolor{green!60!black}{\ding{51}}   & \textcolor{red}{\ding{55}} & \textcolor{red}{\ding{55}} & \textcolor{red}{\ding{55}} & \textcolor{green!60!black}{\ding{51}} & \textcolor{green!60!black}{\ding{51}} & \textcolor{red}{\ding{55}} \\
TritonGym\cite{anonymous2025tritongym} & PyTorch-to-Triton &  \textcolor{green!60!black}{\ding{51}}   & \textcolor{red}{\ding{55}} & \textcolor{red}{\ding{55}} & \textcolor{red}{\ding{55}} & \textcolor{green!60!black}{\ding{51}} & \textcolor{green!60!black}{\ding{51}} & \textcolor{red}{\ding{55}} \\
BabelTower \cite{wen2022babeltower} & C-to-CUDA & \textcolor{green!60!black}{\ding{51}}   & \textcolor{red}{\ding{55}} & \textcolor{red}{\ding{55}} & \textcolor{red}{\ding{55}} & \textcolor{green!60!black}{\ding{51}} & \textcolor{green!60!black}{\ding{51}} & \textcolor{red}{\ding{55}} \\
ComputeEval \cite{compute_eval} & Text-to-CUDA    &  \textcolor{red}{\ding{55}}     &  \textcolor{green!60!black}{\ding{51}} & \textcolor{red}{\ding{55}} & \textcolor{red}{\ding{55}} & \textcolor{green!60!black}{\ding{51}} & \textcolor{red}{\ding{55}} & \textcolor{red}{\ding{55}} \\
\textbf{CUDABench} (\textit{Ours})   & Text-to-CUDA    & \textcolor{green!60!black}{\ding{51}}      & \textcolor{green!60!black}{\ding{51}} & \textcolor{green!60!black}{\ding{51}} & \textcolor{green!60!black}{\ding{51}} & \textcolor{green!60!black}{\ding{51}} & \textcolor{green!60!black}{\ding{51}} & \textcolor{green!60!black}{\ding{51}} \\ 
\bottomrule
\end{tabular}
}
\end{table*}

\section{Related Works}
Developing high-performance CUDA kernels is a time-consuming process that requires substantial expertise in parallel programming, memory hierarchies, and GPU architectures \cite{guide2020cuda}. This challenge has motivated recent efforts to leverage LLMs to automate CUDA kernel generation. To evaluate the ability of LLMs to produce efficient and correct CUDA kernels, KernelBench \cite{ouyang2025kernelbench} introduces a benchmarking framework together with the $\text{fast}_p@k$ metric, which jointly measures functional correctness and performance. As one of the most widely adopted benchmarks in this area, KernelBench has inspired a growing body of work on CUDA kernel generation and optimization \cite{baronio2025kevin, li2025cudal1, dong2025stark, kong2025concur}. However, KernelBench is limited to PyTorch-to-CUDA code translation and is largely restricted to PyTorch ML workloads. Moreover, it does not provide reference CUDA implementations.

ComputeEval \cite{compute_eval} introduces a natural-language-based dataset that broadens the scope of CUDA kernel generation tasks. Nevertheless, as shown in Tab. \ref{tab:compare}, compute-eval lacks reference implementations and relies on relatively small-scale test cases. Crucially, it lacks performance evaluation, limiting its ability to assess practical performance and real-world engineering relevance.

For performance evaluation, most existing benchmarks, as well as the associated kernel generation and optimization methods, primarily rely on execution time as the main metric \cite{ouyang2025kernelbench,wen2022babeltower}. However, execution time is influenced not only by the quality of the generated CUDA kernel but also by the underlying GPU hardware specifications, making direct comparisons across platforms and settings challenging.

To address the above limitations, we introduce CUDABench, a benchmark for evaluating LLMs on text-to-CUDA generation across multiple application domains. CUDABench assesses compilation correctness, functional correctness, and hardware-independent performance using the roofline-based CUDABench-Score. Furthermore, we present a comprehensive evaluation of state-of-the-art LLMs using CUDABench.
\section{Text-to-CUDA Dataset: CUDABench-Set}
Firstly, we construct \textbf{CUDABench-Set}, a text-to-CUDA dataset designed to assess LLM capabilities in CUDA kernel generation. Each benchmark task in CUDABench-Set corresponds to a %single 
CUDA kernel. These tasks are instantiated via prompts that provide detailed textual specifications, including kernel descriptions, data types, hardware, etc. Furthermore, they are supported by a verification pipeline (Sec. \ref{sec:pipeline}), comprising a data generator and a validator. The generator produces randomized test input data and the associated reference output. Subsequently, the validator leverages these inputs to execute the generated kernel and evaluates its correctness by comparing the actual results with the reference output. Moreover, we provide reference CUDA kernel codes for all tasks. On top of these components, CUDABench-Set constructs a \textbf{Breadth-Depth-Difficulty} evaluation space to enable systematic coverage.

\begin{figure*}[t]
\centering
    \begin{subfigure}[b]{0.28\textwidth}
        \centering
        \includegraphics[width=\linewidth]{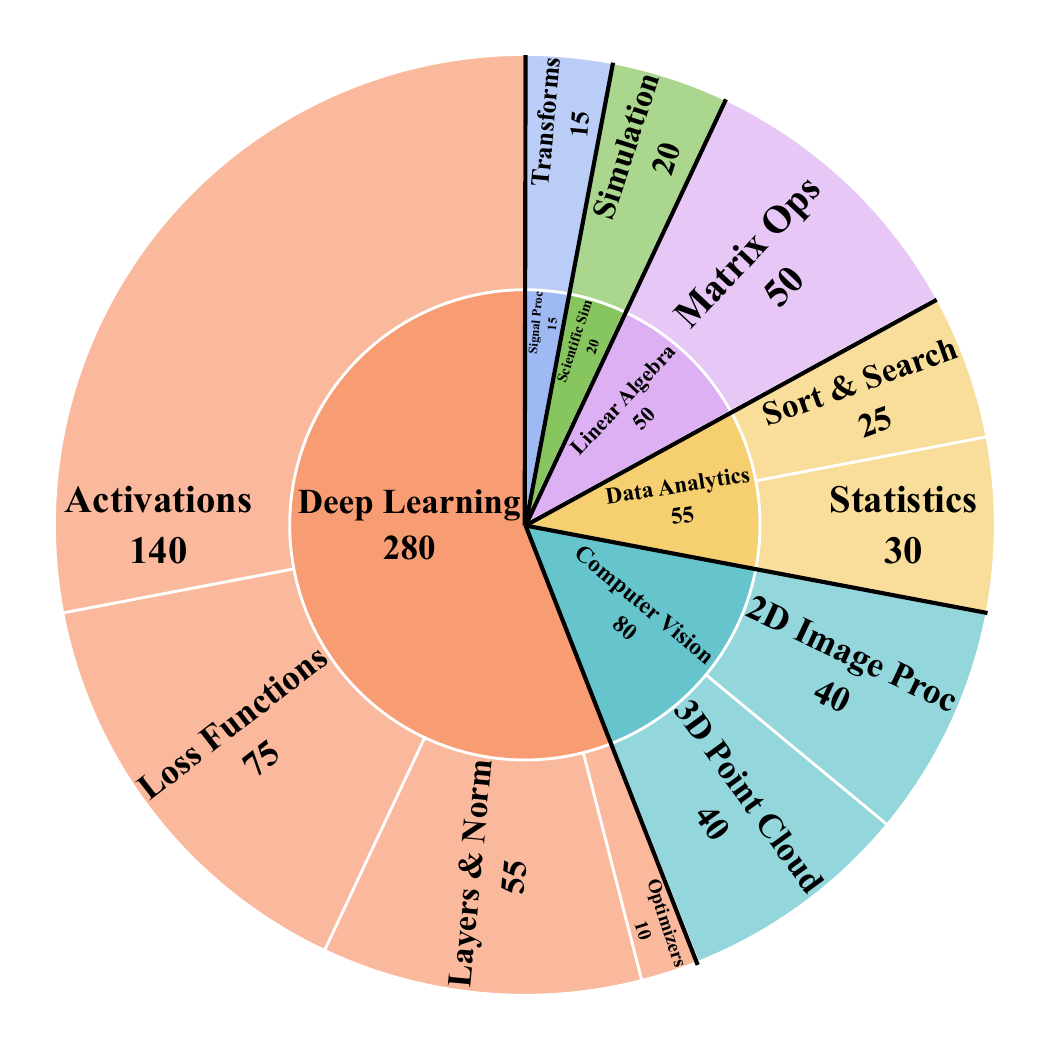}
        \caption{Breadth: Tasks distribution}
        \label{fig:CUDABench-Set}
    \end{subfigure}
    \hfill
    \begin{subfigure}[b]{0.34\textwidth}
        \centering
        \includegraphics[width=\linewidth]{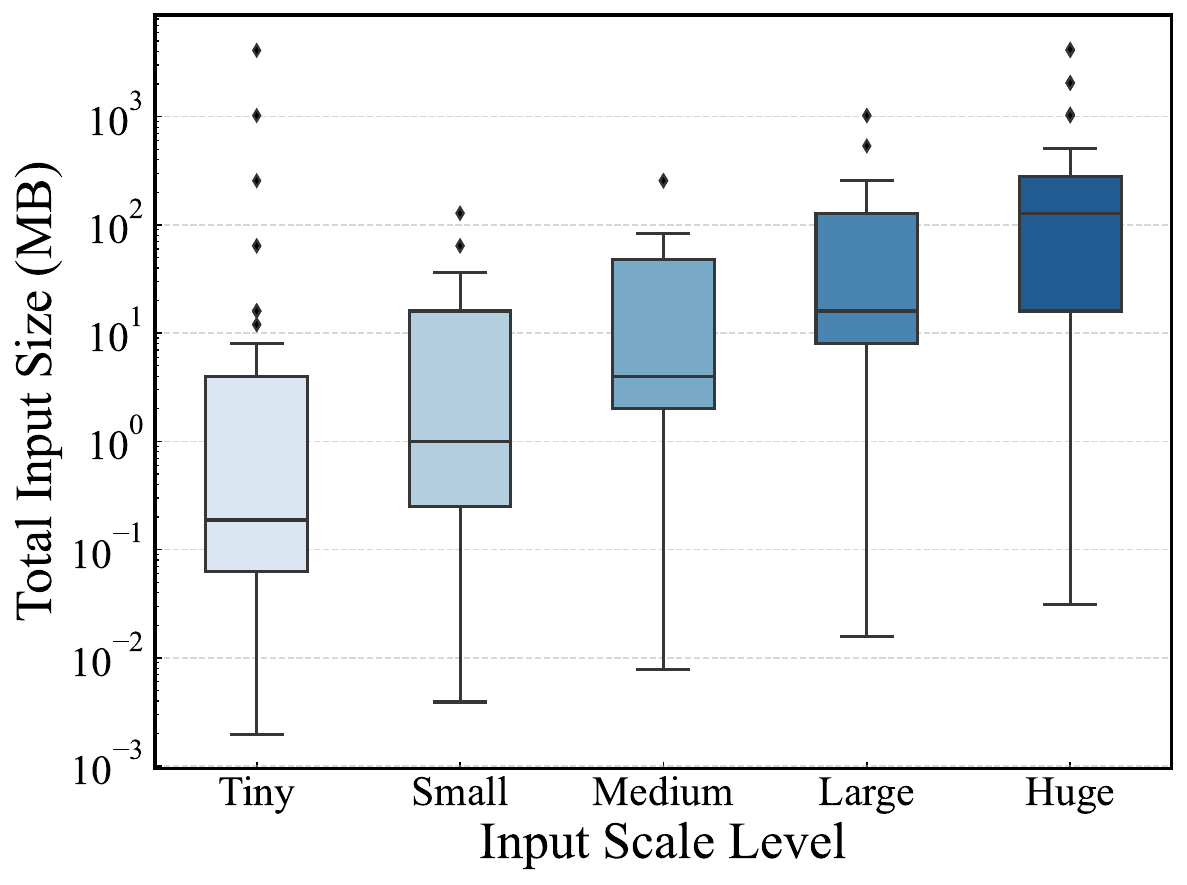}
        \caption{Depth: Input size distribution}
        \label{fig:input_size}
    \end{subfigure}
    \hfill
    \begin{subfigure}[b]{0.34\textwidth}
        \centering
        \includegraphics[width=\linewidth]{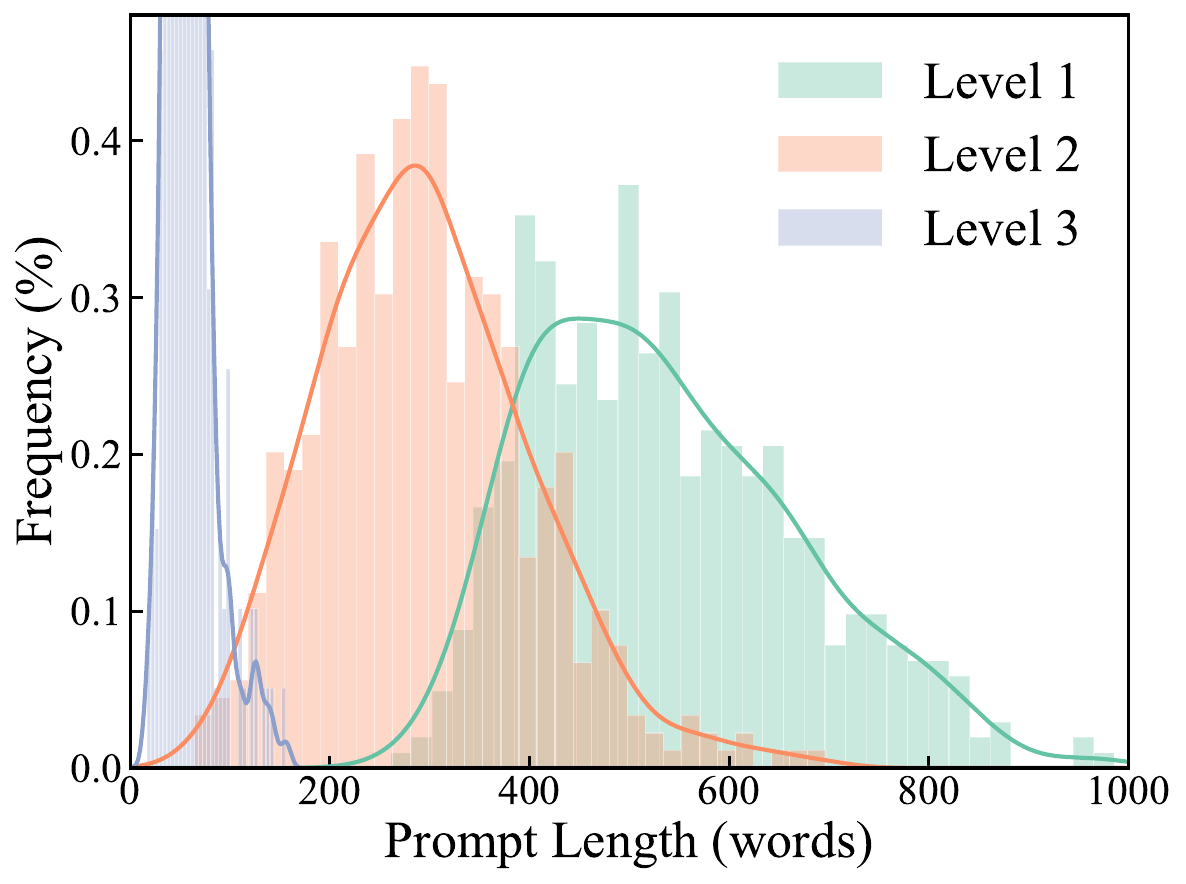}
        \caption{Difficulty: Prompt length}
        \label{fig:prompt_length}
    \end{subfigure}

    \caption{Statistical distribution of CUDABench-Set dimensions. (a) \textbf{Breadth}: Distribution of tasks across domains. (b) \textbf{Depth}: Input size distribution across scale levels. (c) \textbf{Difficulty}: Prompt length distribution across difficulty levels.}
    \label{fig:overall_stats}
\end{figure*}

\paragraph{Breadth.}
CUDABench-Set provides comprehensive coverage of six representative GPU computing domains and a broad range of real-world workloads, including:

\begin{enumerate}[leftmargin=*]
    \item \textbf{Fundamental Linear Algebra}, constituting the bedrock of scientific computing with essential primitives, such as matrix multiplication (GEMM), matrix transposition, element-wise operations, and least squares solvers that underpin higher-level algorithms.
    \item \textbf{Deep Learning Operators}, covering a rich diversity of neural network components, ranging from activation functions (e.g., GELU, Swish) and loss functions (e.g., Cross-Entropy, KL Divergence) to 
    normalization layers (LayerNorm) and optimizers (AdamW).
    \item \textbf{Computer Vision \& Image Processing}, encompassing both traditional 2D image processing techniques (e.g., Gaussian blur, edge detection) and modern 3D point cloud operations (e.g., Ball Query, Furthest Point Sampling) essential for autonomous driving and robotics.
    \item \textbf{Data Analytics}, involving high-throughput statistical and structural primitives, such as sorting, top-$k$ selection, histogramming,  prefix sums (scan), and reduction algorithms used in large-scale data warehousing.
    \item \textbf{Signal Processing}, focusing on time-frequency domain transformations and filtering operations, including Finite Impulse Response (FIR) filtering and Wavelet Transforms (DWT), critical for digital signal analysis.
    \item \textbf{Scientific Simulation \& Finance}, merging high-precision numerical modeling tasks, such as Monte Carlo integration, PDE solvers (e.g., FDTD), and financial derivative pricing models (e.g., Black-Scholes).
\end{enumerate}

All the tasks are constructed from widely used, open-source CUDA codebases, including deep learning operator libraries \cite{paszke2019pytorch,pytorch}, NVIDIA official CUDA Samples~\cite{cudasample}, and other production-level GPU repositories. 
The number and distribution of the tasks is shown in Fig. \ref{fig:CUDABench-Set}.

\paragraph{Depth.}
CUDABench-Set ensures comprehensive performance evaluation through hierarchical test input data design. We define five progressive input scale levels (Tiny, Small, Medium, Large, Huge) for each kernel to mimic a wide spectrum of production-level environments. 
For instance, in the Haar Wavelet Transform task, the input size scales exponentially across the five levels, with 1D signal lengths ranging from 512 to 8,192.
As depicted in the box plot (Fig. \ref{fig:input_size}), the dataset covers a vast range of data sizes, scaling from a few kilobytes in the ``Tiny'' category to over 1 GB in the ``Huge'' category.
By systematically saturating the scale hierarchy and GPU compute units, this design ensures sufficient verification depth, enabling the evaluation of kernel performance under hardware stress.

Note that every input size for each task is paired with a dedicated reference kernel, data generator, and validator, since both algorithmic implementation and validation criteria vary significantly across different scales of input. This granular alignment ensures that the reference output and validation logic are strictly adapted to the computational characteristics of each specific input data size.

\paragraph{Difficulty.}
To explore the LLM capacities from CUDA syntax generation to domain-knowledge retrieval, we construct each benchmark task with prompts of three different difficulty levels that progressively remove contextual and implementation details:

\begin{itemize}[leftmargin=*]
    \item \textbf{Level 1 (Guided Implementation)}, providing the task name, a detailed algorithm description, and explicit CUDA implementation guidelines (e.g., memory hierarchy usage and thread-mapping strategies). This level evaluates LLM capacity to generate architecturally and syntactically correct CUDA code.
    \item \textbf{Level 2 (Algorithmic Specification)}, including the task name and algorithm description, but omits all hardware-specific guidance. At this level, the model must autonomously map scalar algorithmic logic onto a parallel GPU execution model.
    \item \textbf{Level 3 (Concept Retrieval)}, supplying only the task name. This extreme zero-shot setting assesses 
    the ability of LLM
    to internally retrieve the required domain knowledge (e.g., inferring the mathematical formulation of Black–Scholes) and synthesize a correct solution without any external context.
\end{itemize}
 
As shown in Fig.~\ref{fig:prompt_length}, prompts at different levels exhibit distinct length distributions, reflecting the progressive reduction in information density. This hierarchical design allows for a granular evaluation of specific LLM capabilities, ranging from syntax translation to autonomous domain knowledge retrieval.

In summary, CUDABench-Set comprises a total of 1,500 prompts, derived from 500 tasks paired with three difficulty levels. Each task is equipped with a verified reference CUDA kernel, which serves as the ground truth for the accompanying data generator and validator, supporting the subsequent evaluation pipeline.

\section{Text-to-CUDA Metrics: CUDABench-Score}

We first define correctness and performance metrics. Then, we unify these metrics into the \textbf{CUDABench-Score} and detail the \textbf{Generative Verification Pipeline} to automate this end-to-end evaluation framework.

\subsection{Correctness Metrics}
We enforce a strict two-stage filter that a generated kernel is considered valid only if it satisfies:

\begin{enumerate}[leftmargin=*]
    \item \textbf{Compilation Success:} The kernel compiles without error using the standard NVIDIA CUDA Compiler Driver (NVCC)~\cite{nvcc}.
    \item \textbf{Functional Consistency:} The kernel executes successfully across generated test input data, producing results that match the reference output.
\end{enumerate}

Only \textbf{valid} kernels are eligible for the subsequent performance benchmarking in Sec.~\ref{sec:performance_metric}.

\subsection{Performance Metrics}
\label{sec:performance_metric}
Most existing benchmarks~\cite{ouyang2025kernelbench,li2025tritonbench,wen2022babeltower} and optimization~\cite{chen2025cudallm} approaches primarily evaluate CUDA kernels using execution time as the performance indicator \cite{ouyang2025kernelbench,li2025tritonbench}. 
However, execution time is sensitive to hardware configurations, memory bandwidth, system load, and other factors. 
Therefore, we introduce a roofline-based performance metric that evaluates the performance of LLM-generated kernel performance relative to its theoretically attainable performance.

\subsubsection{Performance-Score}
The roofline model is a visual performance model used to provide performance evaluation of a given compute kernel on multi-core, many-core, or accelerator processor architectures~\cite{williams2009roofline}. 
To characterize a kernel on a roofline model, three key quantities must be collected: the execution time, the total number of floating-point operations (FLOPs) performed, and the total volume of data movement. These quantities jointly determine the arithmetic intensity (Arith), which is defined as the ratio of the total number of FLOPs executed by the application to the total number of bytes transferred to support those operations. 

\begin{figure}[t]
\centering
\includegraphics[width=0.95\columnwidth]{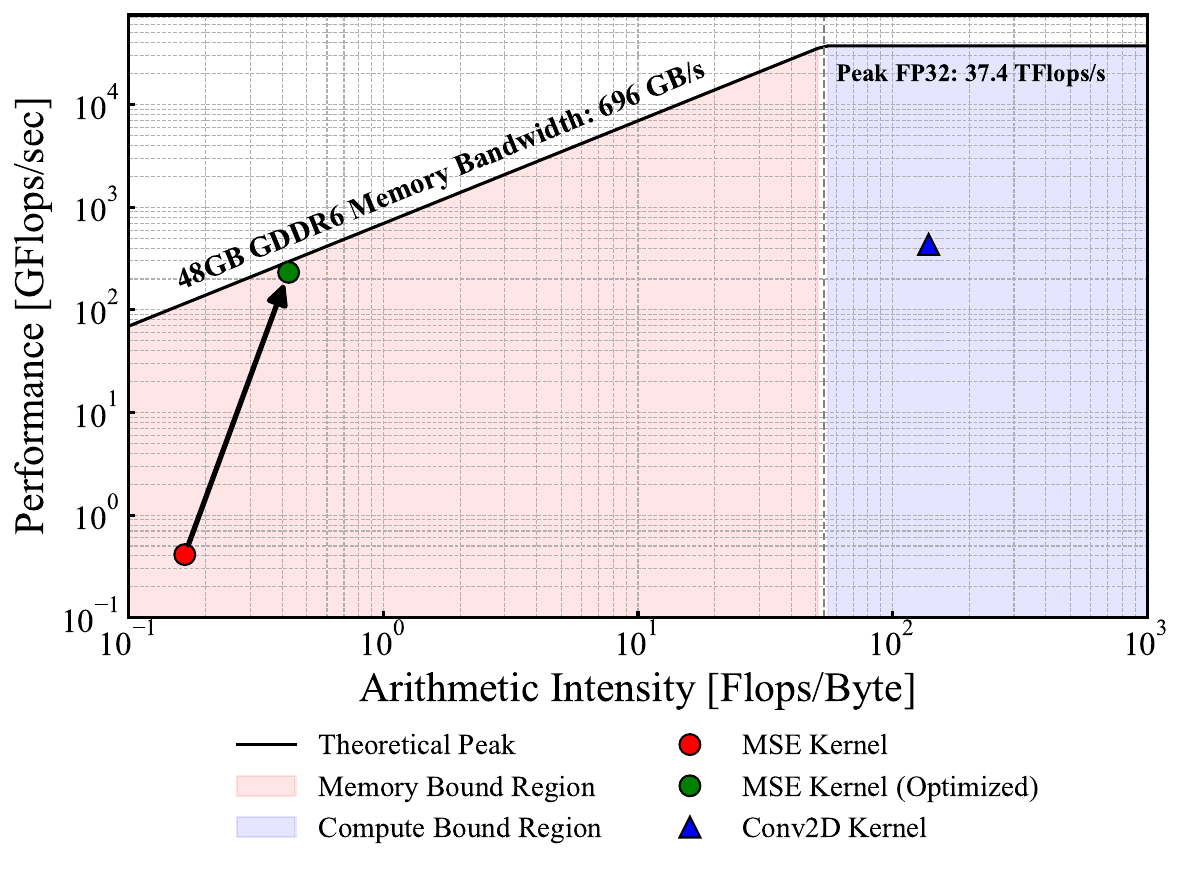}
\caption{The roofline model for an NVIDIA A40 GPU. The $y$-axis is floating-point performance. The $x$-axis is arithmetic intensity. The A40 GPU has peak single-precision floating-point performance of 37.4~TFLOPs/sec and peak memory bandwidth of 696~GB/sec. The chart delineates \textbf{Memory-Bound} (red shaded) and \textbf{Compute-Bound} (blue shaded) regions. It visualizes the performance bottlenecks of an MSE kernel (red dot) and a Conv2D kernel (blue triangle), contrasted with an optimized MSE kernel (green dot) that approaches the theoretical hardware limit.}
\label{fig:roofline model}
\end{figure}

The roofline model defines attainable performance as the theoretical upper bound of performance that a specific kernel can achieve on a given architecture. Formally, this limit is determined by the minimum of the hardware's peak computational throughput and its memory bandwidth capacity, as expressed in Eq.~\eqref{eq:roofline}~\cite{williams2009roofline},
\begin{equation}
\resizebox{\linewidth}{!}{$
\begin{aligned}
\label{eq:roofline}
\text{Attainable GFLOPs/sec}
=\min\{\text{Peak Floating-Point Performance},\\
\text{Peak Memory Bandwidth}\times\text{Arithmetic Intensity}\}.
\end{aligned}
$}
\end{equation}
For instance, we present the roofline model for an NVIDIA A40 GPU in Fig. \ref{fig:roofline model}. The horizontal line shows the peak floating-point performance of the GPU. The actual floating-point performance of a floating-point kernel cannot be higher than this line. The slanted line bounds the maximum floating-point performance that the memory bandwidth of the GPU can support for a given arithmetic intensity. We evaluate three cases: a Mean Squared Error (MSE) kernel (in the Breadth dimension of the Deep Learning Operators domain, and the Depth dimension with an input size of $2^{10}$), a Conv2D kernel (in the Breadth dimension of the Deep Learning Operators domain, and the Depth dimension with inputs of a $(128,128)$ matrix and $(24,24)$ kernel), and an optimized MSE kernel refined by LLMs. As observed in Fig.~\ref{fig:roofline model}, these kernels have different arithmetic intensities and different achieved performance.

By mapping the specific arithmetic intensity of a kernel to Eq.~\eqref{eq:roofline}, we establish its attainable performance as the theoretical hardware limit.
Based on this, we introduce the \textbf{Performance-Score}, 
measuring how closely its achieved performance approaches the theoretical peak performance. Performance-Score is given by:
\begin{equation}\label{eq:perf}
   \text{Performance-Score} = \frac{\text{ Achieved GFLOPs/sec}}{\text{ Attainable GFLOPs/sec}}.
\end{equation}
This formulation provides a normalized evaluation of kernel performance, enabling fair comparison across kernels with fundamentally different performance bottlenecks.

\begin{figure}[t]
    \centering
    \begin{subfigure}[b]{0.48\columnwidth}
        \centering
        \includegraphics[width=\linewidth]{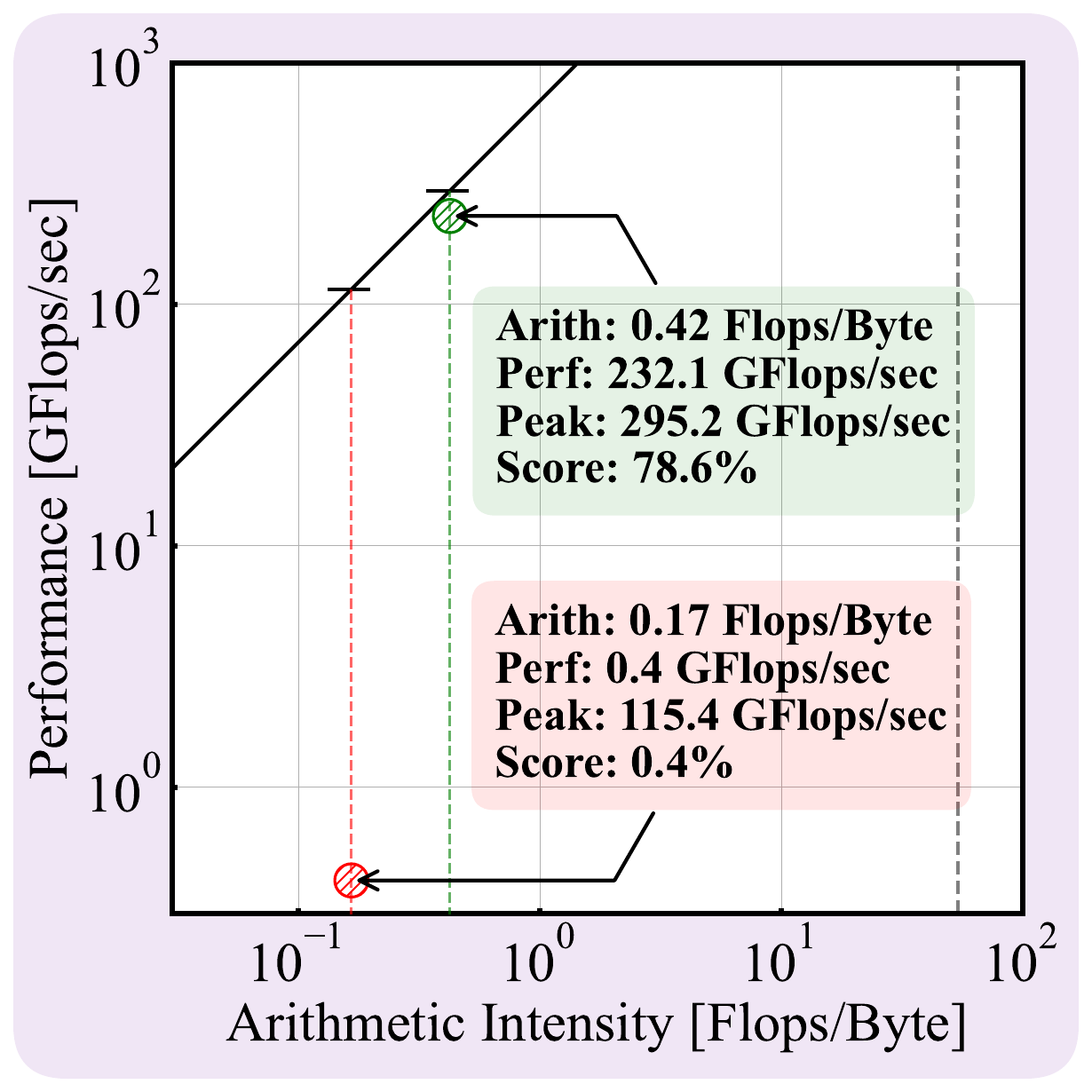}
        \caption{NVIDIA A40 GPU}
        \label{fig:roofline_left}
    \end{subfigure}
    \hfill
    \begin{subfigure}[b]{0.48\columnwidth}
        \centering
        \includegraphics[width=\linewidth]{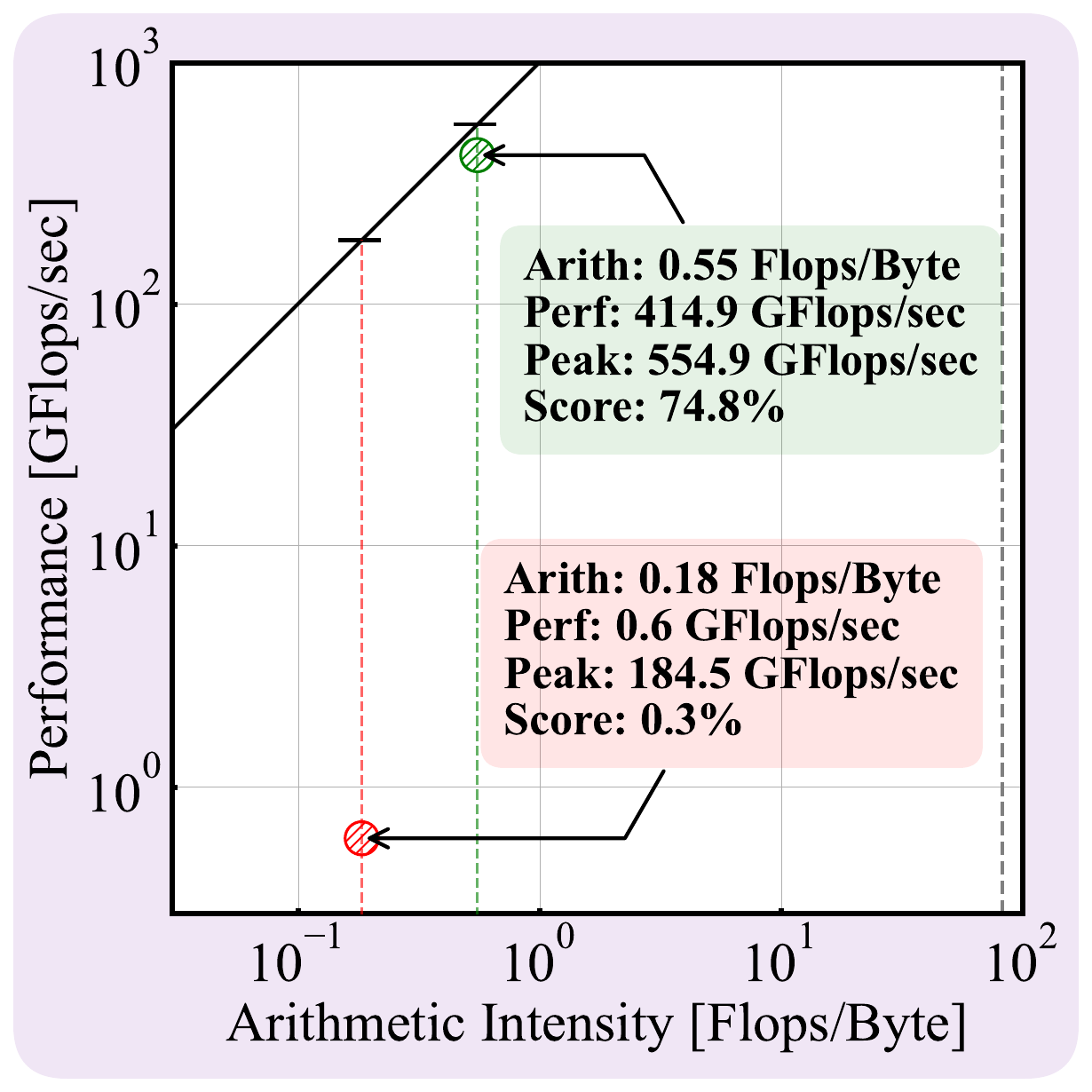}
        \caption{NVIDIA RTX 4090 GPU}
        \label{fig:roofline_right}
    \end{subfigure}
    
    \caption{Comparison of Roofline models for the MSE kernel executed on different GPU. \textbf{Arith} denotes Arithmetic Intensity, \textbf{Perf} refers to the Achieved GFLOPs/sec, \textbf{Peak} represents the Attainable GFLOPs/sec, and \textbf{Score} corresponds to the calculated Performance-Score.}
    \label{fig:two_rooflines}
\end{figure}

As illustrated in Fig.~\ref{fig:two_rooflines}, we zoom out Fig.~\ref{fig:roofline model} and report the  characteristics of MSE kernels executed on NVIDIA A40 GPU and RTX 4090 GPU. 
The achieved performance of the optimized kernel are 295.2 GFLops/sec and 554.9 GFLops/sec on on A40 GPU and RTX 4090 GPU, respectively, while those of the original MSE kernel are 115.4 GFLops/sec and 184.5 GFLops/sec. 
The Performance-Scores of optimized kernel are 78.6\% and 74.8\% on A40 GPU and RTX 4090 GPU, respectively, while those of the original MSE kernel are 0.4\% and 0.3\%.  
The Performance-Score remains relatively consistent,
demonstrating that it effectively normalizes hardware differences, serving as a hardware-independent indicator.

\subsubsection{Memory-Bound \& Compute-Bound}
The roofline model maps the arithmetic intensity of each kernel to its attainable performance, distinguishing whether a workload is limited by memory bandwidth or computational throughput.
For memory-bound kernels, substituting the corresponding term from Eq.~\eqref{eq:roofline} into Eq.~\eqref{eq:perf} yields:
\begin{equation}
\resizebox{0.9\linewidth}{!}{$
\begin{aligned}
    \text{Performance-Score} &=\frac{\text{Achieved GFLOPs/sec}}{\text{Attainable GFLOPs/sec}} \\
    &=\frac{\text{Arith}\times\text{Achieved Memory Bandwidth}}{\text{Arith}\times \text{Peak Memory Bandwidth}} \\
    &=\text{Memory Bandwidth Utilization},
\end{aligned}
$}
\end{equation}
which means the Performance-Score effectively reduces to an evaluation of memory bandwidth utilization. Therefore, the large score improvement indicates that the kernel substantially increases effective memory bandwidth usage, moving much closer to the memory roof. This behavior is expected and desirable, as memory bandwidth is the dominant performance factor for low arithmetic intensity kernels.

Symmetrically, for compute-bound kernels operating to the right of the ridge point, the same metric naturally transitions to measuring compute utilization,
\begin{equation}
\resizebox{0.8\linewidth}{!}{$
\begin{aligned}
    \text{Performance-Score} &=\frac{\text{Achieved GFLOPs/sec}}{\text{Attainable GFLOPs/sec}} \\
    &=\frac{\text{Achieved GFLOPs/sec}}{\text{Peak GFLOPs/sec}} \\
    &=\text{Compute Utilization}. 
\end{aligned}
$}
\end{equation}
In this sense, Performance-Score provides a unified performance metric that adapts to the kernel’s bottleneck regime, quantifying bandwidth utilization for memory-bound workloads and compute efficiency for compute-bound workloads.

\subsection{CUDABench-Score}
\label{sec:CUDABench-Score}
We define \textbf{CUDABench-Score}, a unified metric that integrates compilation correctness, functional consistency, and Performance-Score into a scalar value, which is given by: 
\begin{equation}
\resizebox{0.9\linewidth}{!}{$
\begin{aligned}
    \text{CUDABench-Score} = \frac{1}{N}\sum_{i=0}^N&\left[\right.\mathbbm{1}(\text{Compilation}_i \land \text{Functional}_i)\\
    &\times \text{Performance-Score}_i\left.\right].
    \label{eq:cudabench_score}
\end{aligned}
$}
\end{equation}
where $N$ represents the total number of tasks in the CUDABench-Set, and $i$ denotes the index of the $i$-th task.
Consequently, CUDABench-Score reflects the LLM capability to generate reliable, high-performance CUDA kernels within CUDABench-Set.

\subsection{Generative Verification Pipeline}
\label{sec:pipeline}
To derive CUDABench-Score for CUDABench-Set, we build a generative verification pipeline as part of CUDABench. The pipeline is designed as a fully automated, end-to-end framework that validates and evaluates LLM-generated kernels.
The framework integrates \textbf{Data Generators} and \textbf{Validators} to execute a sequential evaluation pipeline comprising four key stages: (1) test data preparation; (2) kernel compilation; (3) execution-based functional verification; and (4) performance evaluation.

For each task, the data generator produces random input data and corresponding reference outputs. To guarantee robust verification, we ensure the input values are large enough to avoid underflow, making it easier to verify the correctness within the error tolerance. Then the validator compiles and executes the generated code, checking its functional consistency against reference outputs. Kernels that pass correctness verification are subsequently profiled using NVIDIA Nsight Compute \cite{nsys} to measure the execution time, the number of FLOPs, and the volume of data movement, and calculate the CUDABench-Score. Utilizing the real-time generation strategy, the pipeline synthesizes and consumes data without requiring persistent storage. This approach eliminates the heavy disk space occupancy typically associated with large-scale static benchmarks.
\begin{table*}[t]
\centering
\caption{Results from the latest LLMs on CUDABench. Performance comparison of state-of-the-art LLMs across three difficulty levels. We report Pass@1 and Pass@3 rates for compilation correctness, functional consistency, and performance score. Best results are \textbf{bolded}, and second-best are \underline{underlined}. 
{\small Specific model versions evaluated include: 
\nolinkurl{gpt-5.2-2025-12-11}, 
\nolinkurl{gemini-3-flash-preview} (2025-12-17), 
\nolinkurl{claude-sonnet-4-5-20250929}, 
\nolinkurl{deepseek-reasoner} (2025-12-01), 
\nolinkurl{qwen3-max-2025-09-23}, 
\nolinkurl{kimi-k2-thinking} (2025-11-06), 
and \nolinkurl{MiniMax-M2.1} (2025-12-23).}
}
\label{tab:main}
\resizebox{0.95\textwidth}{!}{
\begin{tabular}{@{}llcccccc@{}}
\toprule
\multirow{2}{*}{\textbf{Dataset}} & \multirow{2}{*}{\textbf{Model}} & \multicolumn{3}{c}{\textbf{Pass@1}} & \multicolumn{3}{c}{\textbf{Pass@3}} \\
\cmidrule(lr){3-5} \cmidrule(l){6-8} 
 & & Compilation & Function & Score & Compilation & Function & Score \\ \midrule
\multirow{7}{*}{Level 1} 
& GPT-5.2 (High) \cite{openai2025gpt5}          & 93.4\% & 79.8\% & \textbf{40.9\%} & \underline{99.6\%} & \underline{88.6\%} & \textbf{45.4\%} \\
& Gemini 3 Flash \cite{DeepMind_GeminiFlash_2025}     & \underline{97.6\%} & \underline{83.0\%} & 40.1\% & \textbf{100.0\%} & \underline{88.6\%} & \underline{44.2\%} \\
& Claude 4.5 Sonnet \cite{Anthropic_ClaudeSonnet45_2025} & \textbf{99.8\%} & \textbf{85.8\%} & \underline{40.2\%} & \textbf{100.0\%} & \textbf{90.0\%} & 42.9\%\\
& DeepSeek-V3.2 \cite{liu2025deepseek}          & 96.0\% & 65.2\% & 31.6\% & \underline{99.6\%} & 77.6\% & 38.9\% \\
& Qwen3 Max \cite{yang2025qwen3}                & 93.8\% & 76.8\% & 32.1\% & 98.6\% & 85.2\% & 36.5\% \\
& Kimi K2 Thinking \cite{team2025kimi}          & 82.6\% & 69.8\% & 35.1\% & 95.6\% & 85.6\% & 41.3\% \\
& MiniMax-M2.1 \cite{MiniMax_2_1_2026}          & 87.0\% & 55.2\% & 26.7\% & 98.8\% & 77.0\% & 36.9\% \\ \midrule
\multirow{7}{*}{Level 2} 
& GPT-5.2 (High) \cite{openai2025gpt5}          & 93.4\% & 75.2\% & \underline{37.9\%} & 99.0\% & \underline{85.0\%} & \textbf{44.5\%} \\
& Gemini 3 Flash \cite{DeepMind_GeminiFlash_2025}     & \underline{98.6\%} & \underline{77.6\%} & 37.6\% & \textbf{100.0\%} & 83.2\% & 41.7\% \\
& Claude 4.5 Sonnet \cite{Anthropic_ClaudeSonnet45_2025} & \textbf{99.2\%} & \textbf{82.8\%} & \textbf{39.4\%} & \textbf{100.0\%} & \textbf{86.6\%} & \underline{42.6\%} \\
& DeepSeek-V3.2 \cite{liu2025deepseek}          & 94.2\% & 63.4\% & 31.4\% & 99.6\% & 77.6\% & 37.9\% \\
& Qwen3 Max \cite{yang2025qwen3}                & 92.6\% & 73.8\% & 30.2\% & 98.4\% & 80.4\% & 35.2\% \\
& Kimi K2 Thinking \cite{team2025kimi}          & 64.8\% & 53.8\% & 29.1\% & 84.4\% & 73.0\% & 37.2\% \\
& MiniMax-M2.1 \cite{MiniMax_2_1_2026}          & 89.6\% & 57.6\% & 26.6\% & \underline{99.8\%} & 76.0\% & 38.5\% \\ \midrule
\multirow{7}{*}{Level 3} 
& GPT-5.2 (High) \cite{openai2025gpt5}          & 92.4\% & \underline{60.8\%} & \textbf{29.8\%} & 99.4\% & \textbf{70.6\%} & \textbf{35.5\%} \\
& Gemini 3 Flash \cite{DeepMind_GeminiFlash_2025}     & \underline{96.8\%} & 59.4\% & \underline{28.4\%} & \textbf{100.0\%} & 68.6\% & \underline{33.6\%} \\
& Claude 4.5 Sonnet \cite{Anthropic_ClaudeSonnet45_2025} & \textbf{99.0\%} & \textbf{63.2\%} & \underline{28.4\%} & \textbf{100.0\%} & \underline{69.8\%} & 32.3\% \\
& DeepSeek-V3.2 \cite{liu2025deepseek}          & 96.4\% & 49.8\% & 22.5\% & \textbf{100.0\%} & 61.4\% & 28.7\% \\
& Qwen3 Max \cite{yang2025qwen3}                & 92.0\% & 50.6\% & 22.5\% & 99.2\% & 61.0\% & 26.8\% \\
& Kimi K2 Thinking \cite{team2025kimi}          & 76.4\% & 49.0\% & 23.5\% & 92.4\% & 63.2\% & 30.5\% \\
& MiniMax-M2.1 \cite{MiniMax_2_1_2026}          & 86.6\% & 42.6\% & 20.1\% & \underline{99.6\%} & 59.8\% & 28.8\% \\ \bottomrule
\end{tabular}
}
\end{table*}

\section{Experiment}

\subsection{Experimental Setup} \label{sec:setup}
\paragraph{Model.}
We evaluate CUDABench using the latest LLMs, including several models with specialized reasoning modes. Specifically, we include: 
\textit{GPT-5.2 (High)}~\cite{openai2025gpt5}, 
\textit{Claude 4.5 Sonnet}~\cite{Anthropic_ClaudeSonnet45_2025}, 
\textit{Gemini 3 Flash}~\cite{Google_Gemini3_2025,DeepMind_GeminiFlash_2025}, 
\textit{DeepSeek-V3.2} \cite{liu2025deepseek}, 
\textit{Kimi K2 Thinking} \cite{team2025kimi}, 
\textit{Qwen3-Max}~\cite{yang2025qwen3}, and 
\textit{MiniMax-M2.1}~\cite{MiniMax_2_1_2026}. 
All models are accessed via their official APIs to guarantee that the evaluation reflects their most current capabilities.

\paragraph{Hardware.} We primarily conduct experiments on the NVIDIA A40 GPU, based on the Ampere architecture \cite{ampere_arch}. The A40 GPU is equipped with 48 GB of GDDR6 memory providing 696 GB/s of bandwidth and consists of 84 Streaming Multiprocessors (SMs) delivering a peak FP32 throughput of 37.4 TFLOPS. To evaluate performance generalization, we provide supplementary results using the NVIDIA GeForce RTX 4090 GPU (Ada Lovelace architecture \cite{ada_arch}). While the RTX 4090 GPU has a smaller memory capacity (24 GB GDDR6X), it offers significantly higher computational density with 128 SMs, achieving a peak FP32 performance of 82.6 TFLOPS and a memory bandwidth of 1,008 GB/s.

\paragraph{Metrics.}
We evaluate the generated kernels using \textbf{Compilation Success}, \textbf{Functional Correctness}, and \textbf{Performance-Score}. We report results using \textbf{Pass@\textit{k}} ($k=\{1, 3\}$)~\cite{chen2021code}, which estimates the probability that at least one valid or high-performance CUDA kernel is generated within $k$ sampling attempts.

\subsection{Evaluation Results}

Tab. \ref{tab:main} presents the comprehensive evaluation results of different LLMs on \textbf{CUDABench}. We report results across three difficulty levels. 
In terms of overall performance, Claude 4.5 Sonnet demonstrates superior capability in correctness, consistently achieving the highest metrics in both compilation and functional consistency. Specifically, under the Level 1 (Pass@1) setting, it attains a near-perfect compilation rate of 99.8\% and a leading functional accuracy of 85.8\%. Conversely, regarding kernel performance, GPT-5.2 (High) has a slight advantage, securing the top CUDABench-Score of 40.9\% at Level 1 and maintaining this lead across higher difficulty levels. While other model also delivers competitive results, a deeper examination of these metrics reveals critical limitations shared across all models.
By analyzing the metric discrepancies across different models and levels, we derive the following findings:

\paragraph{Finding 1: CUDABench poses significant challenges for LLMs.} 
The tested LLMs exhibit distinguishable performance on CUDABench. 
Taking GPT-5.2 as an example, under the Pass@1 setting, its functional accuracy is only 79.8\% at Level~1 and drops significantly to 60.8\% at Level~3, indicating increasing difficulty as kernel complexity grows.

This performance stands in contrast to the strong results previously reported on general-purpose code generation benchmarks such as HumanEval~\cite{chen2021evaluating}, where even earlier models (e.g., GPT-5) achieved accuracies above 90\%. 
This discrepancy suggests that strong general code generation capabilities do not directly generalize to the domain of CUDA kernel generation, which requires specialized knowledge of parallel programming, memory hierarchies, and performance constraints.
Also, we observe a notable improvement in Pass@3 compared to Pass@1. 
This indicates that LLMs are capable of generating correct CUDA kernels given multiple attempts, suggesting substantial headroom for future improvements through better prompting, sampling strategies, or domain-specific training.

\begin{figure}[t]
\centering
\includegraphics[width=0.95\columnwidth]{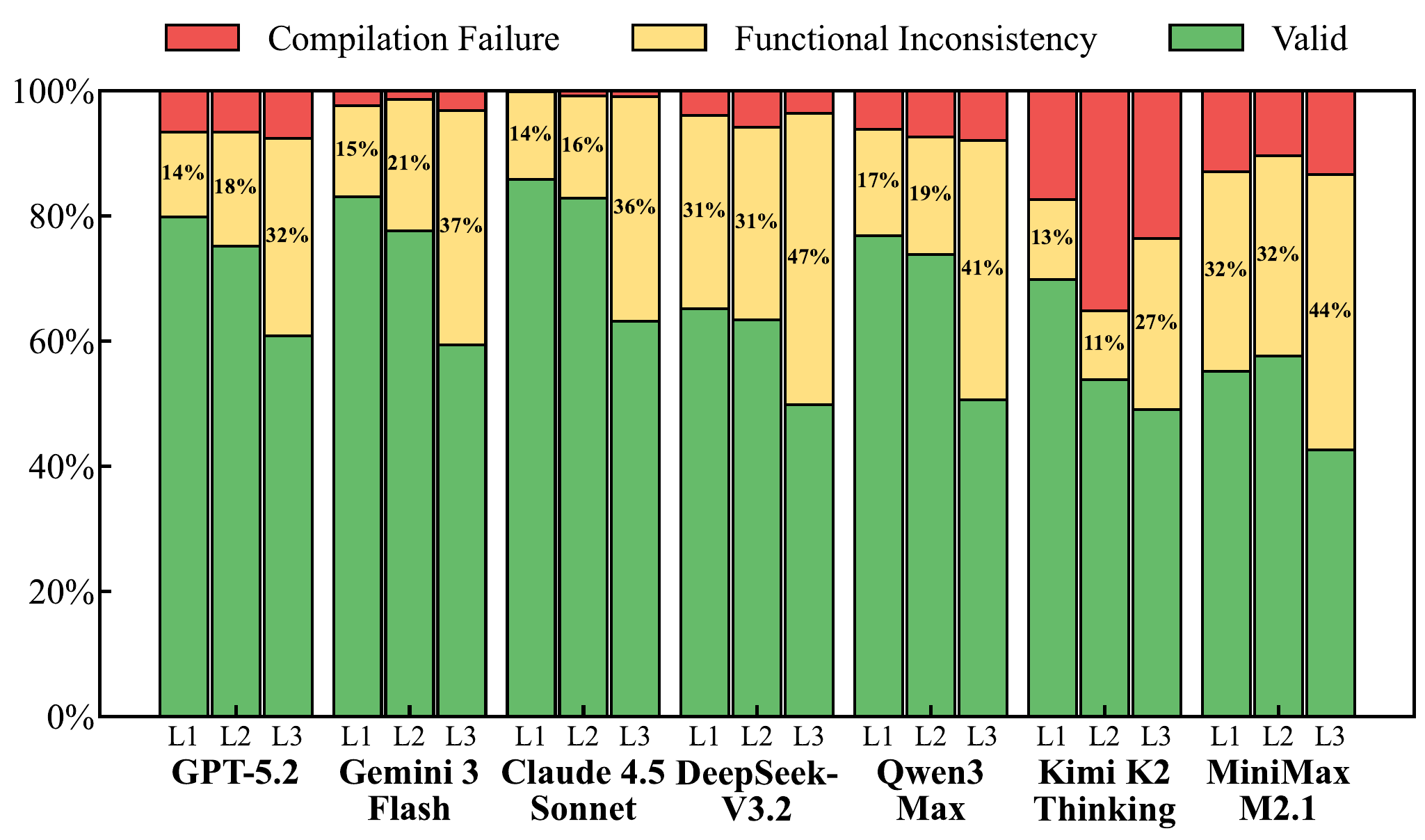}
\caption{The proportion of Compilation Failure (Red), Functional Inconsistency (Yellow), and Valid (Green) kernel for each model. Results are broken down by difficulty levels (Pass@1).}
\label{fig:error}
\end{figure}

\paragraph{Finding 2: Generated kernels achieve high compilation rates but low functional correctness.} We observe a substantial divergence between syntactic validity and semantic accuracy. As illustrated in Fig. \ref{fig:error}, functional errors are the primary source of failure, accounting for the majority of incorrect outputs and substantially exceeding compilation errors across nearly all evaluated models. While model like Claude 4.5 Sonnet achieve a perfect 100\% compilation rate across all difficulty levels in Tab. \ref{tab:main} , their ability to generate functionally accurate code declines sharply. This trend is most pronounced in DeepSeek-V3.2 at Level 3, where functional errors account for approximately 47\% of generations, whereas compilation faults remain 3.6\%. This distribution indicates that modern LLMs have effectively mastered CUDA syntax and API conventions, yet they frequently generate superficially valid but logically flawed code, failing to handle complex semantics such as thread synchronization and memory boundary conditions.

\paragraph{Finding 3: LLMs lack domain-specific knowledge and CUDA implementation expertise.}

In the Level 3 evaluation, without algorithmic explanations and CUDA implementation hints, we provide models only with the task names and I/O formats. Consequently, we observe a sharp performance decline on Level 3 across all LLMs compared to Levels 1 and 2. This degradation highlights a critical retrieval generation bottleneck: without explicit external guidance, models struggle to retrieve the correct algorithmic primitives and implementation patterns from their internal weights. This evidence strongly suggests that current LLMs suffer from a fundamental scarcity of domain-specific knowledge and lack the intrinsic expertise required for CUDA implementation.

\begin{figure}[t]
\centering
\includegraphics[width=0.95\columnwidth]{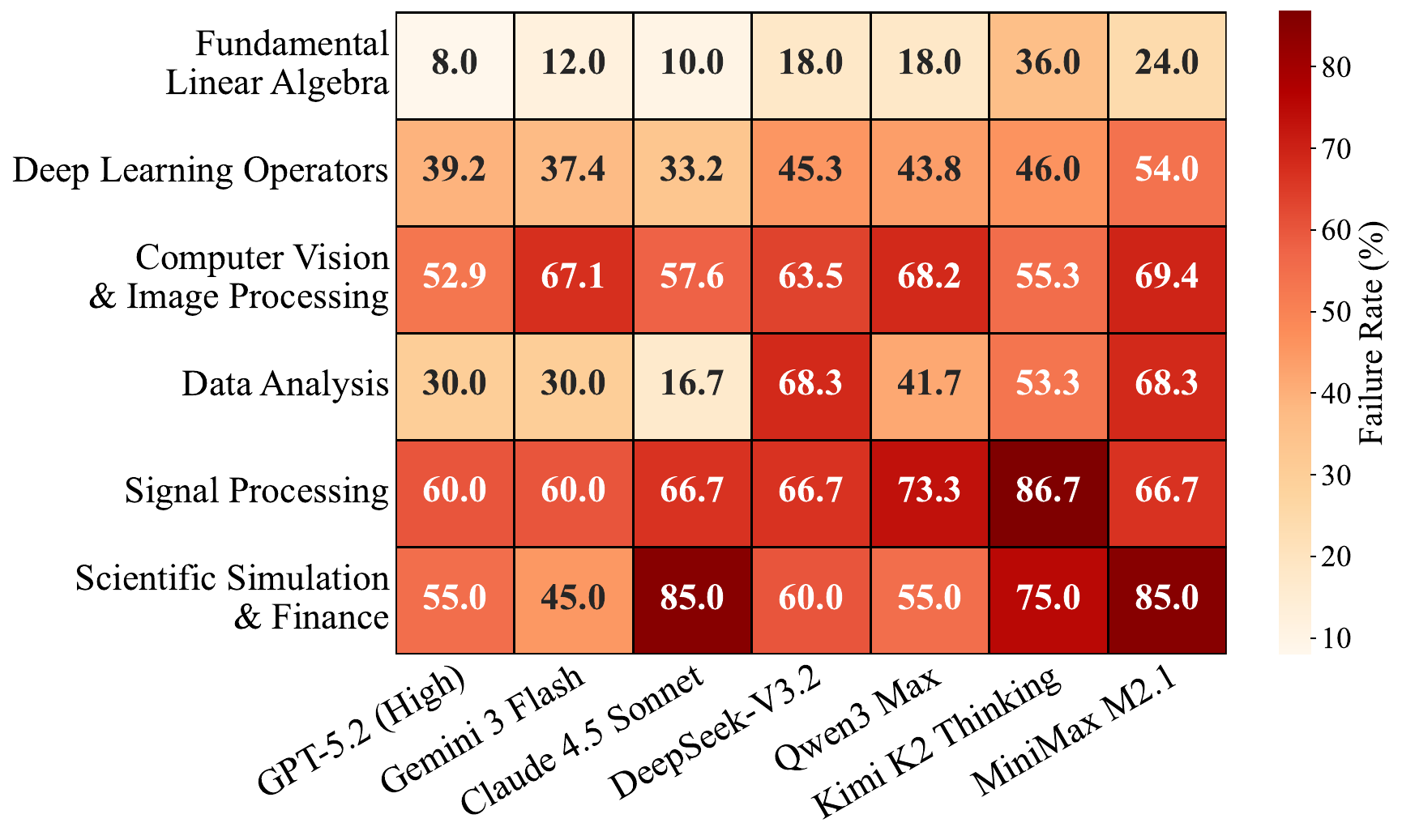}
\caption{Analysis of failure rates on Level 3 tasks (Pass@1), accounting for kernels that either compilation failure or function inconsistency.}
\label{fig:heatmap}
\end{figure}

As shown in Fig. \ref{fig:heatmap}, further analysis of failure rates on Level 3, across multiple domains, supports this finding through a more granular analysis. While models exhibit relative proficiency in foundational tasks like ``Linear Algebra'' (e.g., GPT-5.2 shows a low failure rate of 8.0\%), they struggle significantly in more comprehensive domains. Specifically, in the ``Scientific Simulation \& Finance'' category, failure rates surge, with Claude 4.5 Sonnet reaching a staggering 85.0\% failure rate. This domain-level disparity underscores that LLMs' internal knowledge is skewed towards general mathematical concepts, while remaining critically deficient in the complex, niche algorithms required for signal processing and scientific computing.

\paragraph{Finding 4: Generated kernels demonstrate low execution performance on practice hardware.} To evaluate execution performance, we utilize the CUDABench-Score, a roofline-based metric that quantifies the proximity of a kernel's performance to the hardware's theoretical peak (Sec. \ref{sec:CUDABench-Score}). The results in Tab. \ref{tab:main} reveal a common inefficiency across all evaluated models. Even at Level 1, the performance of top-tier models is capped at a mediocre level, leaving approximately 60\% of the GPU's computational resources unexploited. Notably, this performance bottleneck is systemic rather than model-specific, as evidenced by the negligible performance gap among leading models. For instance, GPT-5.2, Claude 4.5 Sonnet, and Gemini 3 Flash achieve nearly identical performance scores of 40.9\%, 40.2\%, and 40.1\%, respectively. This uniform challenge suggests that while current LLMs are capable of generating syntactically correct code, they cannot universally implement critical hardware-aware optimizations, resulting in implementations that fail to fully leverage the massive capacity of modern GPUs.

\subsection{Supplementary Results on RTX 4090 GPU}

\begin{table}[]
\centering
\caption{Supplementary results on NVIDIA RTX 4090 GPU.}
\label{tab:sup}
\resizebox{0.95\columnwidth}{!}{
\begin{tabular}{@{}llcccc@{}}
\toprule
\multirow{2}{*}{\textbf{Dataset}} & \multirow{2}{*}{\textbf{Model}}& \multicolumn{2}{c}{\textbf{Pass@1}} & \multicolumn{2}{c}{\textbf{Pass@3}} \\ \cmidrule(l){3-4} \cmidrule(l){5-6} 
                         &                  & Function      & Score      & Function      & Score      \\ \midrule
\multirow{4}{*}{Level 1} & DeepSeek-V3.2    & 64.0\% & 28.2\% & 78.9\%& 34.1\%     \\
                         & Qwen3 Max        & 76.8\% & 28.0\% & 85.9\% & 31.1\%     \\
                         & Kimi K2 Thinking & 51.5\% & 25.5\% & 69.7\% & 34.4\%     \\
                         & MiniMax-M2.1     & 56.2\% & 24.0\% & 79.4\% & 33.9\%\\\midrule
\multirow{4}{*}{Level 2} & DeepSeek-V3.2    & 62.8\% & 27.1\% & 75.9\%& 33.4\%  \\
                         & Qwen3 Max        & 75.6\% & 27.6\% & 82.6\% & 30.6\%\\
                         & Kimi K2 Thinking & 56.4\% & 26.9\% & 73.5\% & 34.6\% \\
                         & MiniMax-M2.1     & 55.3\% & 22.9\% & 76.4\% & 32.8\%\\\midrule
\multirow{4}{*}{Level 3} & DeepSeek-V3.2    & 51.1\% & 20.0\% & 63.0\%& 26.2\%\\
                         & Qwen3 Max        & 54.8\% & 19.7\% & 61.0\% & 21.6\% \\
                         & Kimi K2 Thinking & 50.1\% & 23.7\% & 63.0\% & 28.9\% \\
                         & MiniMax-M2.1     & 42.6\% & 18.0\% & 60.0\% & 26.1\% \\ \bottomrule
\end{tabular}
}
\end{table}

To evaluate the generalization of our metrics, we extended our experiments to the NVIDIA GeForce RTX 4090 GPU. As detailed in Sec.~\ref{sec:setup}, the RTX 4090 GPU possesses significantly higher theoretical peak performance and memory bandwidth compared to the NVIDIA A40 GPU used in the main experiments. For this supplementary analysis, we specifically selected a subset of four LLMs that demonstrated suboptimal performance on the NVIDIA A40 GPU.

Despite this substantial disparity in hardware specifications, the evaluation results demonstrate a remarkable consistency in the CUDABench-Score across platforms. As shown in the Tab. \ref{tab:sup}, the CUDABench-Score for identical models remain stable. For instance, at Level 1, Qwen3 Max achieves a CUDABench-Score of 28.0\% on the RTX 4090 GPU, which is highly comparable to 32.1\% on the A40 GPU.

This stability indicates that the CUDABench-Score effectively normalizes hardware differences. 
Consequently, the metric serves as a robust, hardware-independent indicator, allowing for fair benchmarking of LLM capabilities regardless of the underlying accelerator.

\section{Conclusion}
We introduced CUDABench, a comprehensive and hardware-independent benchmark that establishes a Breadth-Depth-Difficulty evaluation space and a novel roofline-based performance metric CUDABench-Score to systematically evaluate text-to-CUDA generation. Our extensive evaluation of state-of-the-art LLMs reveals that despite achieving high compilation success rates, current models struggle with a significant disparity in functional correctness, suffer from a fundamental scarcity of domain-specific knowledge, and fail to fully leverage GPU hardware resources. CUDABench laid the foundation for research on CUDA kernel generation.

% \section*{Impact Statement}

% This paper presents work whose goal is to advance the field of 
% Machine Learning. There are many potential societal consequences 
% of our work, none which we feel must be specifically highlighted here.


\bibliography{reference}
\bibliographystyle{icml2026}

%%%%%%%%%%%%%%%%%%%%%%%%%%%%%%%%%%%%%%%%%%%%%%%%%%%%%%%%%%%%%%%%%%%%%%%%%%%%%%%
%%%%%%%%%%%%%%%%%%%%%%%%%%%%%%%%%%%%%%%%%%%%%%%%%%%%%%%%%%%%%%%%%%%%%%%%%%%%%%%
% APPENDIX
%%%%%%%%%%%%%%%%%%%%%%%%%%%%%%%%%%%%%%%%%%%%%%%%%%%%%%%%%%%%%%%%%%%%%%%%%%%%%%%
%%%%%%%%%%%%%%%%%%%%%%%%%%%%%%%%%%%%%%%%%%%%%%%%%%%%%%%%%%%%%%%%%%%%%%%%%%%%%%%

\newpage
\appendix
\onecolumn

\section*{Example}

% Examples of three difficulty levels prompts.
Here shows an example prompt for the Haar Wavelet Transform kernel (in the Breadth dimension of Signal Processing domain, and the Depth dimension with an input size of 8,192). 

% \begin{figure}[H]
% \centering
% \begin{footnotesize}
\resizebox{\linewidth}{!}{
\begin{tcolorbox}[title=\textbf{Level 1}, colback=blue!5, colframe=blue!60!black, width=\linewidth]
Implement a CUDA kernel for the Haar wavelet transform (DWT\_Haar). The input is a signal tensor of 8192 float32 values. The kernel must recursively decompose the signal into detail and approximation coefficients. Each thread block processes 1024 elements using 512 threads. The computation involves pairwise operations: detail coefficients are calculated as $(x - y) * 1/\sqrt{2}$ and approximation coefficients as $(x + y) * 1/\sqrt{2}$. Detail coefficients from all decomposition levels must be stored in the 8192-length output tensor, with level-1 details placed in the second half. The final approximation coefficient per block must be stored in an 8-length output tensor. Constraints: fixed input size (8192), 8 blocks, and 512 threads per block.
\end{tcolorbox}
}

\resizebox{\linewidth}{!}{
\begin{tcolorbox}[title=\textbf{Level 2}, colback=blue!5, colframe=blue!60!black]
Compute a Haar wavelet transform on an 8192-element signal. Divide the signal into 8 blocks. For each block, recursively apply pairwise operations: details = $(x - y)/\sqrt{2}$ and approximations = $(x + y)/\sqrt{2}$. Store all detail coefficients in an 8192-output array (level-1 details in indices 4096-8191). Store the final approximation per block in an 8-element array.
\end{tcolorbox}
}

\resizebox{\linewidth}{!}{
\begin{tcolorbox}[title=\textbf{Level 3}, colback=blue!5, colframe=blue!60!black]
Compute the Haar wavelet transform for an 8192-element signal, generating detail coefficients and block-wise approximations.
\end{tcolorbox}
}
% \caption{Examples of three difficulty levels prompts.}
% \label{fig:prompt_example}
% \end{footnotesize}
% \end{figure}

% \newpage
% \appendix
% \onecolumn

% You can have as much text here as you want. The main body must be at most $8$
% pages long. For the final version, one more page can be added. If you want, you
% can use an appendix like this one.

% The $\mathtt{\backslash onecolumn}$ command above can be kept in place if you
% prefer a one-column appendix, or can be removed if you prefer a two-column
% appendix.  Apart from this possible change, the style (font size, spacing,
% margins, page numbering, etc.) should be kept the same as the main body.
%%%%%%%%%%%%%%%%%%%%%%%%%%%%%%%%%%%%%%%%%%%%%%%%%%%%%%%%%%%%%%%%%%%%%%%%%%%%%%%
%%%%%%%%%%%%%%%%%%%%%%%%%%%%%%%%%%%%%%%%%%%%%%%%%%%%%%%%%%%%%%%%%%%%%%%%%%%%%%%

\end{document}